\documentclass{article}

\usepackage{microtype}
\usepackage{graphicx}
\usepackage{subfigure}
\usepackage{booktabs}
\usepackage[accepted]{icml2021}

\usepackage{comment}
\usepackage{amsmath,amssymb} 
\usepackage{color}
\usepackage{epsfig}
\usepackage{acro}
\usepackage{multirow}
\makeatletter
\DeclareRobustCommand\onedot{\futurelet\@let@token\@onedot}
\def\@onedot{\ifx\@let@token.\else.\null\fi\xspace}

\DeclareAcronym{WSI}{
short=WSI,
long=whole slide image,
foreign-plural={}
}
\DeclareAcronym{FNAC}{
short=FNAC,
long=Cytology of Fine Needle Aspirates,
foreign-plural={}
}
\DeclareAcronym{FNAB}{
short=FNAB,
long=Fine Needle Aspiration Biopsy,
foreign-plural={}
}
\DeclareAcronym{GEP}{
short=GEP,
long=Gene Expression Profile,
foreign-plural={}
}
\DeclareAcronym{UM}{
short=UM,
long=Uveal Melanoma,
foreign-plural={}
}
\DeclareAcronym{ROI}{
short=ROI,
long=Region of Interest,
foreign-plural={}
}
\DeclareAcronym{mAP}{
short=mAP,
long=Mean Average Precision,
foreign-plural={}
}
\DeclareAcronym{IoU}{
short=IoU,
long=Intersection-Over-Union,
foreign-plural={}
}
\DeclareAcronym{HCI}{
short=HCI,
long=Human Computer Interaction,
foreign-plural={}
}
\DeclareAcronym{SVM}{
short=SVM,
long=Support Vector Machine,
foreign-plural={}
}

\newcommand{\new}[1]{\textcolor{black}{#1}}

\def\eg{\emph{e.\,g. }} 
\def\ie{\emph{i.\,e. }}

\begin{document}
%
\icmltitlerunning{An Interpretable Algorithm for UM subtyping from cytology images}
%

\twocolumn[
\icmltitle{An Interpretable Algorithm for Uveal Melanoma Subtyping from Whole Slide Cytology Images}




\begin{icmlauthorlist}
\icmlauthor{Haomin Chen}{cs}
\icmlauthor{T. Y. Alvin Liu}{med}
\icmlauthor{Catalina Gomez}{cs}
\icmlauthor{Zelia Correa}{med}
\icmlauthor{Mathias Unberath}{cs}
\end{icmlauthorlist}

\icmlaffiliation{cs}{Department of Computer Science, Johns Hopkins University, MD, USA}
\icmlaffiliation{med}{Wilmer Eye Institute, School of Medicine, Johns Hopkins University, MD, USA}

\icmlcorrespondingauthor{Haomin Chen}{hchen135@jhu.edu}
\icmlcorrespondingauthor{Mathias Unberath}{unberath@jhu.edu}

\icmlkeywords{Machine learning, Digital pathology, Cancer, Oncology, Instance segmentation, Object Detection, UMAP}

\vskip 0.3in
]
\printAffiliationsAndNotice{}


%
%
\begin{abstract}
Algorithmic decision support is rapidly becoming a staple of personalized medicine, especially for high-stakes recommendations in which access to certain information can drastically alter the course of treatment, and thus, patient outcome; a prominent example is radiomics for cancer subtyping. Because in these scenarios the stakes are high, it is desirable for decision systems to not only provide recommendations but supply transparent reasoning in support thereof. For learning-based systems, this can be achieved through an interpretable design of the inference pipeline. Herein we describe an automated yet interpretable system for uveal melanoma subtyping with digital cytology images from fine needle aspiration biopsies. Our method embeds every automatically segmented cell of a candidate cytology image as a point in a 2D manifold defined by many representative slides, which enables reasoning about the cell-level composition of the tissue sample, paving the way for interpretable subtyping of the biopsy. Finally, a rule-based slide-level classification algorithm is trained on the partitions of the circularly distorted 2D manifold. This process results in a simple rule set that is evaluated automatically but highly transparent for human verification. On our in house cytology dataset of 88 uveal melanoma patients, the proposed method achieves an accuracy of $87.5\,\%$ that compares favorably to all competing approaches, including deep ``black box'' models. The method comes with a user interface to facilitate interaction with cell-level content, which may offer additional insights for pathological assessment.


\end{abstract}

\section{Introduction}

\begin{figure*}[t]
   \centering
   \includegraphics[width=\linewidth]{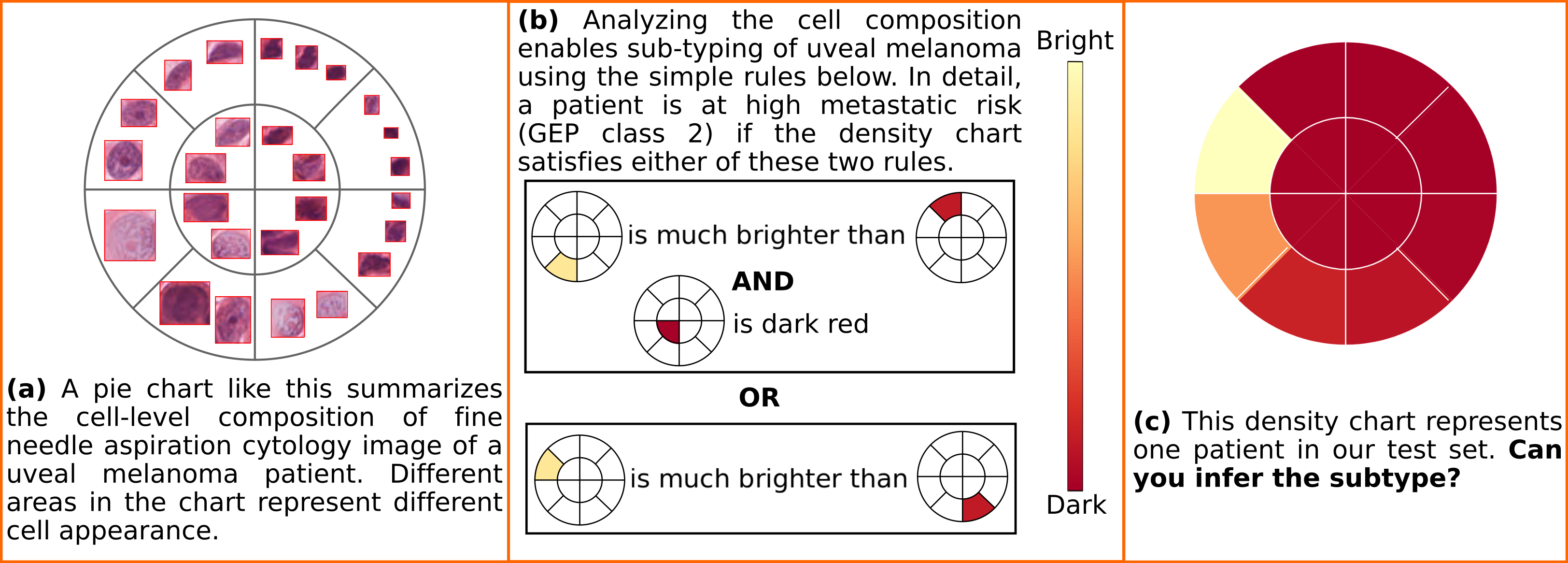}
   \caption{An overview of the automatic interpretable algorithm for uveal melanoma subtyping from cytology images. The algorithm consists of cell instance segmentation to extract cell appearance information that is used to cluster cells of similar appearance in a circular space. Based on a coarse partitioning of the embedding space, which we refer to as pie chart, shown in (a), we find simple rule sets (b) that enable uveal melanoma subtyping, which otherwise, requires gene analysis. A pie chart of a representative patient is shown in (c) - the patient is at high metastatic risk (GEP class 2).}
   \label{fig:head}
\end{figure*}

\ac{UM} is the most common primary intraocular malignancy in adults~\cite{SINGH20111881}. As standard care for \ac{UM}, \ac{FNAB} is often performed to confirm the diagnosis and enable \ac{UM} prognostication. To this end, a molecular test, \ac{GEP}, is performed and microscopic Cytology of Fine Needle Aspirates images are created from the biopsy. According to a recent study, there exist two subtypes in \ac{UM} that can be identified based on its \ac{GEP}: The first subtype exhibits low metastatic risk, while the second subtype has been linked to high metastatic risk. There is a stark contrast in long-term survival between the two classes: the 92-month survival probability in class 1 patients is 95\%, versus 31\% in class 2 patients~\cite{onken2004gene}. It is evident that access to \ac{UM} subtype information is critical for proper management of patients by providing appropriate recommendation for metastasis surveillance. However, even after 10 years of development, \ac{GEP} is still only available in the United States. The technique is also expensive and has a long turn around time.
A more accessible test for \ac{UM} subtyping is, therefore, highly desirable.

There is increasing evidence that the underlying genetic profile affects cancer growth on multiple scales. Radiomics, for example, exploits this observation to develop imaging-derived biomarkers that are informative for prognosis~\cite{Grossmann2017DefiningTB}. In the particular case of \ac{UM} prognostication,
there is huge potential in using imaging-derived biomarkers to determine \ac{GEP} subtype and metastatic risk directly from cytology slides. While it is impossible even for highly trained pathologists to derive this information from cytology images, learning-based algorithms that discover associations between intensity patterns in cytology images and \ac{GEP} subtype are promising~\cite{liu2020gene,chen2020interactive}. However, as ``black box'' models that perform a super-human task, these algorithms do not offer insights beyond the final recommendation to the human decision makers, which has been linked to automation bias and over-trust or dis-trust in such systems~\cite{nourani2020role,gaube2021ai}. A more transparent algorithm design may enable humans to better calibrate their trust in the recommendation, which would be an important feat for high-stakes decision making.

\begin{figure*}[t]
   \centering
   \includegraphics[width=\linewidth]{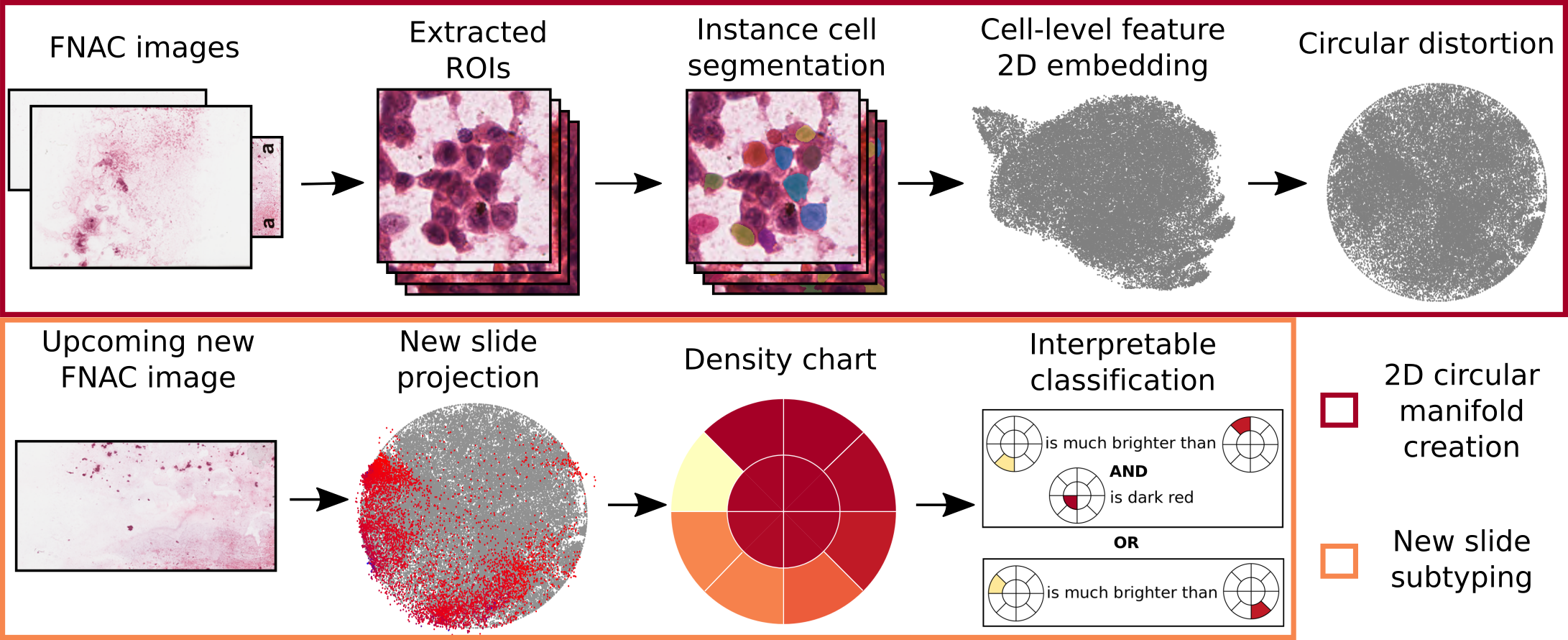}
   \caption{System overview of the proposed method.
   Cell-level features are obtained by aggregation over instance cell segmentation masks and then embedded into a 2D space. Several slides are embedded in this way to create a representative cell appearance space, and the 2D embedding space is subsequently distorted to a circle. For every other \ac{WSI}, cells representations are extracted and projected with the same embedding process into the circular space, such that one density chart is generated for every slide. Finally, we find an interpretable rule set to classify \ac{UM} biopsies based on the density charts.}
   \label{fig:overview}
\end{figure*}

In this paper, we develop an automatic system for interpretable \ac{UM} subtype classification from cytology images. The method is based on the idea that biopsy samples of the two \ac{UM} subtypes should differ in overall cell composition. Thus, an algorithm that enables high level, rule-based reasoning on the cell composition of the sample, would be interpretable and could easily be verified by human users. To create this algorithm, we have developed automated methods for cell instance segmentation in cytology images from weak supervision, techniques to aggregate and represent whole slide-level cell appearance information in an intuitive embedding space, and rule-based classification algorithms to infer \ac{UM} subtype from this representation.   
Figure~\ref{fig:head} provides a concise overview of the user-facing side of the method.  
Because subtyping now amounts to evaluating simple instructions, the rule-based system is highly transparent and offers insight into whether a specific sample ``barely" or ``strongly" obeys the rules, which may enable calibration of trust in the system.
The present manuscript details the technical developments that were necessary to devise this system, and our future work will focus on human factors, including trust and over-reliance issues, in this high-stakes, high-knowledge imbalance scenario. 

\section{Related work}
The high resolution and complexity of \acp{WSI} make cell-level annotations difficult or impossible to obtain, which is why many of the annotated datasets are limited to slide-level labels that correspond to the overall diagnosis. However, standard automatic cancer subtyping and analysis in \acp{WSI} is based on multiple small regions extracted from slides, that then need to be aggregated to a single prediction on the slide level. These methods include majority voting, coarse-to-fine techniques~\cite{liu2017detecting,hou2016patch,xu2017large,zhang2020piloting}, 
and multiple instance learning approaches~\cite{chikontwe2020multiple,campanella2019clinical}. While most of the deep learning approaches in \acp{WSI} analysis consider black box models, recent works attempt to introduce features that enhance model understanding by mimicking the decision process of pathologists. For instance, content-based histopathological image retrieval~\cite{peng2019multi,hegde2019similar} contrasts a query image with a large database to determine the search results with more similar histological features. 
Making intelligent systems interpretable is another frontier in developing trustworthy medical decision support applications~\cite{rudin2019stop}. In contrast to explainable models that rely on post hoc analysis, interpretable models aim to explain the reasoning behind a prediction. In a histopathologic context, patch-based regions visualizations introduced in~\cite{pirovano2020improving} display features related to tumor tissue, in addition to providing slide-level heatmaps that improve \ac{WSI} classification. 
Different from previous methods that attempt to provide human-meaningful visualizations, either from learned representations or image retrieval, our method is interpretable by nature.

\section{Method}
Given high-quality \acp{ROI} extracted from cytology images, we create an interpretable system to analyze \ac{UM} biopsy cytology and reveal \ac{GEP} subtype based on overall cell composition of the sample. Our learning-based system comprises of three parts: 1) instance-level cell segmentation, 2) cell feature embedding, and 3) interpretable classification. We note that the specifics of each component may easily be replaced by other techniques since aspects pertaining to the exact method choice are not the main focus of our work. In the remainder of this chapter, we describe a cost-efficient way of weakly labeling our dataset to enable supervised learning of the cell segmentation network (Section~\ref{sec:segmentation}). Cell-level features are then generated and embedded into a 2D space for further classification as described in Section~\ref{sec:mapping}. Finally, we define an interpretable classification model within the 2D space to distinguish \ac{UM} \ac{GEP} classes by cell composition (Section~\ref{sec:interpretable_classification}). The system overview is shown in Figure~\ref{fig:overview}.



\subsection{Considerations around interpretability}
\new{
Before we introduce the technical details of the proposed method, we first frame our model in the current interpretability definition. As proposed in~\cite{Murdoch2019}, interpretable machine learning
is defined to be the use of machine-learning models for the extraction of \textit{relevant} knowledge about domain relationships contained in data. Knowledge is considered to be relevant if it provides insight for a particular audience into a chosen domain problem. Our proposed method aims to provide an interpretable model for \ac{UM} \ac{GEP} classification to pathologists, who already possess substantial expertise in reading cytology slides. Indeed, clinicopathologic features of \ac{UM} tumors, \eg epithelioid cell type and aggressiveness of cancer cells, can be readily estimated from cytology images and have been associated with worse patient prognosis and a higher incidence of metastatic disease~\cite{worley2007transcriptomic}. These risk factors are widely used by pathologists, however, their accuracy to predict metastatic potential has been shown to be limited~\cite{schopper2016clinical}. Our interpretability mechanism is motivated by the fact that clinicopathologic features of \ac{UM} tumors, e.\,g., cell appearance, are \textit{relevant} knowledge for pathologists to predict \ac{UM} metastatic risk. Instead of building interpretable models with the clinicopathologic information manually extracted by pathologists themselves, our system automatically extracts clinicopathologic features (cell appearance features) and summarizes the cell appearance distribution in a 2-dimensional space, which is further classified by a simple and interpretable rule set. 
While the algorithm was developed in close collaboration with ophthalmic oncologists and pathologists, this manuscript is limited to describing the proposed system and characterizing its performance and does not empirically demonstrate its interpretability on a larger user group. Doing so will remain subject of future work. 
}

\subsection{Instance cell segmentation}
\label{sec:segmentation}
\begin{figure}[t]
   \centering
   \includegraphics[width=\linewidth]{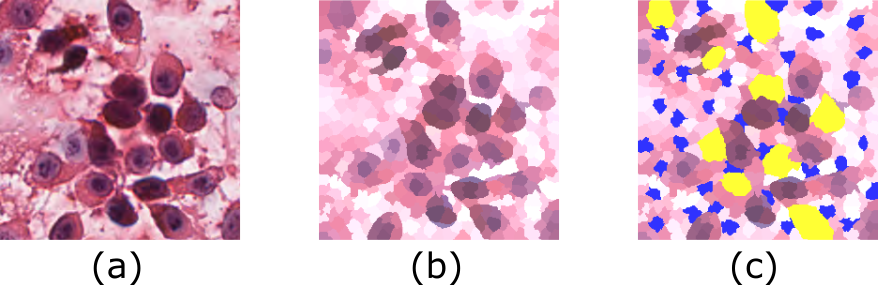}
   \caption{The \ac{ROI} annotation procedure. (a) the extracted high-quality \ac{ROI}; (b) the generated super-pixels.; (c) the annotations on super-pixels. Yellow and blue regions are annotated super-pixels for cancer cells and background, respectively.}
   \label{fig:annotation}
\end{figure}

There exist no cell annotations for the high-quality \acp{ROI} that are automatically extracted from cytology images using the method described in~\cite{chen2020interactive}. However, instance cell segmentation is essential for further cell-level analysis. Thus, we prepare annotations on a small sub-set with minimal manual labor to enable supervised training of an instance segmentation network. Figure~\ref{fig:annotation} presents the annotation procedure. In detail, we randomly select 500 \acp{ROI} from the 131k pool and partially annotate super-pixels generated by SLIC~\cite{SLIC} to reduce the annotation workload. We group all super-pixels within any annotated cells to generate instance-level annotations. We trained the instance segmentation network YOLACT~\cite{YOLACT} on the annotated \acp{ROI}, by converting annotated super-pixels into pixel level annotations. \new{We chose YOLACT because it can be easily modified to enable training on partially annotated data. Unlike two-stage segmentation networks, \eg Mask RCNN~\cite{he2017mask}, which first detects candidate regions to then classify and segment these regions in the second stage, YOLACT breaks the instance segmentation into two parallel tasks: (1) generating a dictionary of non-local prototype masks over the entire image, and (2) predicting a set of linear combination coefficients per instance.} We compute all loss functions, \eg semantic segmentation loss, only in annotated areas. All \acp{ROI} are finally tested to extract cells. 
\subsection{Cell-level feature embedding}
\label{sec:mapping}
Previously, pathologists have attempted to quantify different cell components, such as nuclear size and nucleolar size, to predict the behavior of tumors. 
Our approach is similar to this process as it extracts network feature representations of cells, which we assume contain information about cell appearance. Cell-level features $F_c$ for cell $c$ are then extracted from the entire feature map $F$ using the segmentation mask $M_c$ with masked average pooling:
\begin{align}
    F_c = \text{Avg}(F[M_c])\,,
\end{align}
\new{where $F$ is the output of the backbone network architecture.}

To prepare subtype classification based on cell composition, and improve classification performance, 
we embed all cell-level features in 20 \ac{GEP} class 1 slides to create a 2D embedding space with UMAP~\cite{UMAP}. All other slides are then embedded into that space, to represent the respective cell composition. We expect slides of distinct \ac{GEP} classes have different cell composition, and thus distribution in the 2D embedding space. \new{The embedding space is created using slides of one \ac{GEP} class only to potentially maximize the difference of \ac{GEP} class 1 and class 2 representation in the embedding space.}
\begin{figure*}[t]
   \centering
   \includegraphics[width=\linewidth]{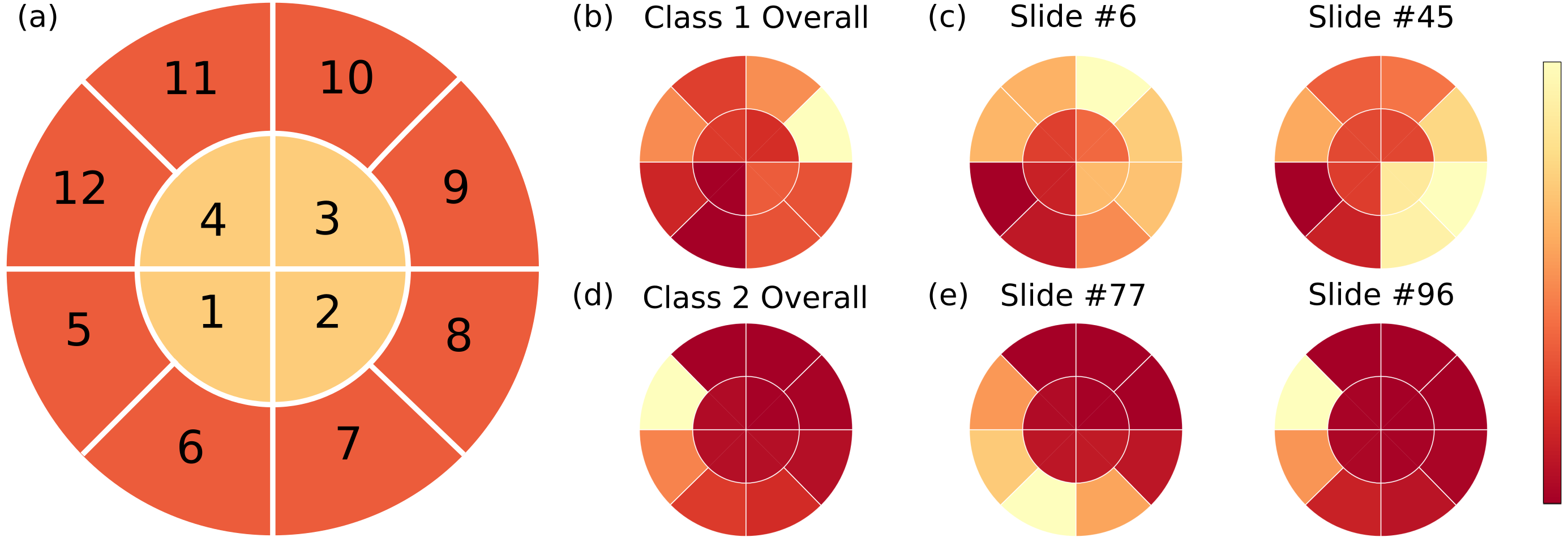}
   \caption{(a) The definition of spatial partitioning and density charts in the distorted 2D embedding space. (b) The density chart of all cells in \ac{GEP} class 1; (c) Two density chart examples of \ac{GEP} class 1 slides; (d) The density chart of all cells in \ac{GEP} class 2; (e) Two density chart examples of \ac{GEP} class 2 slides. Examples in (c) and (e) are all correctly predicted samples by using the rule set defined in Equation~\ref{eqn:rule_set}.}
   \label{fig:interpretable_classification}
\end{figure*}
\subsection{Interpretable Classification}
\label{sec:interpretable_classification}
Based on our hypothesis that slide-level cell composition, and thus distributions in the 2D cell appearance embedding space, should be different between \ac{GEP} classes, we devise an interpretable algorithm that reasons based on these representations. Direct comparisons between distributions, \eg chi-square test~\cite{pearson1900x} and Kolmogorov-Smirnov tests~\cite{kolmogorov1933sulla}, are complicated and not usually interpretable. Instead, we partition the embedding space and analyze the region densities. 
Because cells with similar appearance, thus similar features, are close to each other in the embedding space, the density of each region represents the portion of cells with a specific kind of cell appearance in the slide. 
To make it easier to define the spatial partitioning of the embedding space, we first distort the space into a unit circle. We treat the center of gravity of all embedded cells as the origin. Then, we normalize to unity the scale of all embedded cells in every degree of angle in polar coordinate, so that the whole embedding space is distorted to a unit circle. \new{Parameters in circular distortion are determined simultaneously with the embedding generation and are fixed when embedding new slides.} Finally, we divide the unit circle equally into 12 regions, as shown in Figure~\ref{fig:interpretable_classification}.  Since we posit that each \ac{GEP} class will have different densities in distinct regions, in addition to the individual densities of these regions ($D_i$), we define the relative densities ($D_i/D_j$) as input variables for classification. Finally, an interpretable bayesian rule set algorithm~\cite{wang2017bayesian} takes all these 78 input variables (12 values ($D_i$), and 66 relations ($D_i/D_j$)) for \ac{GEP} classification.  

\new{Different from logistic regression (which is only interpretable in low dimensional problem settings because humans can handle at most $7 \pm 2$ cognitive entities at once~\cite{miller202011,cowan2010magical}), 
the rule set algorithm is not limited by the number of input variables. The number of arguments in each rule can be controlled, as simple as determining the largest allowed depth of a tree. In addition, it is different from a random forest (which uses majority vote, and is thus not interpretable) since here, the predicted output is determined once the sample obeys at least one rule in the rule set.}


\section{Experiments}
We demonstrate that our proposed interpretable learning pipeline does not compromise on performance of \ac{UM} subtyping when compared to deep black box models. 
We also demonstrate an additional, interactive tool for expert review of cell level composition through interaction with the 2D embedding space. By simply clicking on areas of interest in the density charts, users may retrieve and visually inspect cells that are representative of the appearance in that specific embedding location. 
Details are shown in Figure~\ref{fig:GUI}.

\subsection{Experimental setting}
\textbf{Dataset:} The dataset we use includes 100 cytology samples from 88 uveal melanoma patients. \new{To the best of our knowledge, this is the largest dataset on \ac{UM} cytology. The dataset contains 50 slides from 43 patients with \ac{GEP} class 1 and 50 slides from 45 patients with \ac{GEP} class 2.} The cellular aspirates obtained from cytology of each tumor were submitted for cytology and \ac{GEP} testing. The cytology specimen was flushed on a standard pathology glass slide, smeared, and stained with hematoxylin and eosin. The specimen submitted for \ac{GEP} was flushed into a tube containing extraction buffer and submitted for DecisionDx-UM testing. Whole slide scanning was performed for each cytology slide at a magnification of 40x. Automatic \ac{ROI} extraction is performed using~\cite{chen2020interactive}, resulting in a total of $131,816$ high-quality \acp{ROI} across all slides.

\textbf{Implementation details: }
Super-pixel algorithm SLIC~\cite{SLIC} is implemented following~\cite{fastSLIC}, where the number of components is $400$, and the Euclidean distance ratio is $1$. On average, each of the 500 randomly sampled \ac{ROI} for manual annotation has 9 cells and 38 background super-pixels annotated. The number of prototypes in YOLACT is doubled to 64 to potentially segment more cells within every \ac{ROI}. The segmentation model is optimized using Adam~\cite{kingma2014adam} with a learning rate of $10^{-5}$ and 4000 iterations with a batch size of $1$. We train the model on $450$ annotated \acp{ROI} and validate on the other $50$ \acp{ROI}. We empirically split the circular embedding space into 12 partitions, as shown in Figure~\ref{fig:interpretable_classification}, which in internal development was found to yield the best performance compared to other split approaches. 
All cells that map outside the circular embedding space are projected to the nearest region. For the interpretable classification, we use $80\%$ of the projected slides in both class 1 and class 2 for training \new{(64)} and the other $20\%$ for testing \new{(16)}. \new{The rule set algorithm is trained with simulated annealing procedure as described in~\cite{wang2017bayesian}.} The maximal length of each rule in the rule set is set to 2 to preserve its intelligibility.

\subsection{Cell Segmentation Performance}
\begin{figure}[t]
   \centering
   \includegraphics[width=\linewidth]{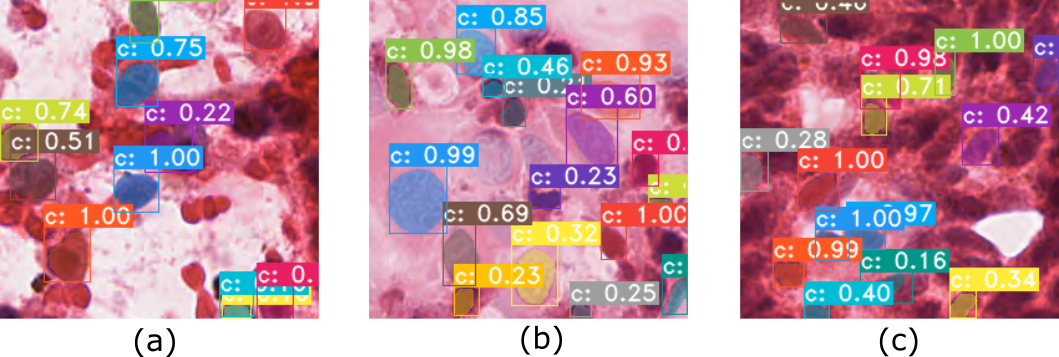}
   \caption{Examples of segmentation results. The segmentation network is able to (a) separate cancer cells (purple, large) from blood cells (red, small); (b) segment cells with all sizes, but (c) misses some ambiguous cells. The numbers within the boxes correspond to confidence scores.}
   \label{fig:segmentation_result}
\end{figure}

We use \ac{mAP} as the main evaluation metric for cell segmentation performance. The \ac{mAP} is about $70\%$ when \ac{IoU} is larger than $50\%$, which indicates that the segmentation process catches a fairly good number of cancer cells. However, \ac{mAP} is low with high \ac{IoU} threshold, because of the low quality of super-pixel-based annotations on the cells' boundary. Table~\ref{tab:segmentation} and Figure~\ref{fig:segmentation_result} present both quantitative and qualitative results, respectively. The algorithm can easily tell apart cancer cells from blood cells, while some cancer cells with ambiguous boundaries are missed. We attribute this to the low quality of super-pixels for these cells during annotation. As a result, cells with ambiguous boundaries are usually skipped in annotation if more clear cells exist in the same \ac{ROI}. Because there exist numerous cells in each slide, missing some cells at random will not significantly impact the overall cell composition\new{, and further, the classification performance}. 

\begin{figure*}[t]
   \centering
   \includegraphics[width=\linewidth]{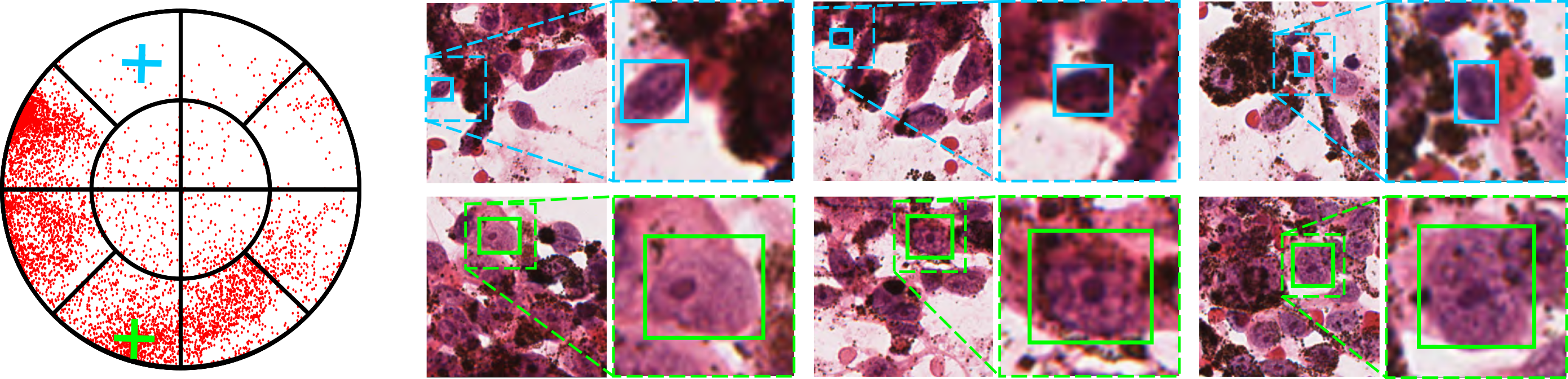}
   \caption{The additional tool for embedding space interaction. For every slide embedding chart, users can click any area of interest, \ie the green/blue cross location. Several closest cells in the embedding space are visualized in native and scaled resolution.}
   \label{fig:GUI}
\end{figure*}

\subsection{\ac{UM} subtype classification}
We compare our proposed method with a previously proposed deep black box model~\cite{coudray2018classification,liu2020gene} evaluated on the same dataset, which classifies \ac{UM} subtype directly from \acp{ROI}. In~\cite{coudray2018classification}, slide-level subtype prediction is obtained by simply averaging class predictions for all corresponding \acp{ROI}. Both, the black box and our proposed method have \new{the same backbone network architecture (ResNet-50~\cite{He2015}) and} the same training and testing split for a fair comparison. We find that the accuracy performance of our method (\textbf{87.5\%}) compares favorably to the black box approach following~\cite{coudray2018classification} (83.3\%), and more importantly, is interpretable based on the following rule set over the density chart ($D_i$), and thus, over the cell appearance composition of the whole slide:
\begin{equation}
\begin{split}
    \label{eqn:rule_set}
    D_{6} / D_{11} > 1.5 &\text{ AND } D_{1} < 0.07 \\
    &\text{OR}\\
    D_{7} / D&_{12} \leq 0.4 
\end{split}
\end{equation}
There only exist $3$ arguments in the rule set, which makes algorithmic recommendations transparent and verifiable, while enabling users to understand overall cell composition. This rule set was visually represented already in Figure~\ref{fig:head}.

\begin{table}[t]
\centering
\caption{mAP for segmentation boxes and masks with different \ac{IoU} threshold.}
\begin{center}
\begin{tabular}{ |l|ccccc| } 

 \hline
 IoU & $0.5$ & $0.6$ & $0.7$ & $0.8$  & $0.9$ \\
 \hline
 box&  $70.67\%$ & $64.41\%$  & $49.20\%$ & $27.52\%$  & $3.24\%$ \\ 
 mask& $69.30\%$ & $64.72\%$  & $53.07\%$ & $33.91\%$  & $2.49\%$ \\
 \hline
\end{tabular}
\end{center}
\label{tab:segmentation}
\end{table}

\begin{table*}[t]
\centering
\caption{Ablation study of interpretable classification with different methods and an ensemble technique. LR refers to logistic regression. Rule Set (class $k$, $k={1,2}$) refers to results using the embedding created from class $k$ slides.}
\begin{center}
\begin{tabular}{ |l|c|c|c|c| } 
 \hline
 &\multicolumn{2}{c|}{w/o Ensemble}  & \multicolumn{2}{c|}{w/ Ensemble} \\
 \cline{2-5}
 & Accuracy & \# of rules & Accuracy & \# of rules\\
 \hline
 LR                 & $67.50 \pm 5.56\% $ & N/A             & $75.14 \pm 9.00\%$ & N/A \\ 
 \hline
 SVM                & $83.00 \pm 6.37\% $ & N/A             & $82.07 \pm 8.23\%$ & N/A \\
 \hline
 ANN                & $82.86 \pm 8.33\% $ & N/A             & $83.71 \pm 10.15\%$& N/A \\
 \hline
 Rule Set (class 1)           & \multirow{1}{*}{\boldmath$86.36 \pm 10.25\%$} & \multirow{1}{*}{$2.28 \pm 0.57$} & \multirow{1}{*}{\boldmath$87.50 \pm 9.56\%$}& \multirow{1}{*}{$2.11 \pm 0.37$}\\
 \hline
  Rule Set (class 2)           & \multirow{1}{*}{$81.93 \pm 8.02\%$} & \multirow{1}{*}{$2.06 \pm 0.49$} & \multirow{1}{*}{$84.33 \pm 10.68\%$}& \multirow{1}{*}{$1.96 \pm 0.31$}\\
 \hline
\end{tabular}
\end{center}
\label{tab:classification_methods}
\end{table*}

\subsection{Ablation Study}
\label{sec:ablation_study}
We conduct an ablation study of the rule-based interpretable classification to benchmark its performance against other classification methods, \ie logistic regression, \ac{SVM} and Artificial Neural Network (ANN). We also compare different embeddings, by creating the initial UMAP embedding space with either, \ac{GEP} class 1 or \ac{GEP} class 2 slides. 
After the embedding space creation, only 80 slides remain to train and evaluate the classification models. Therefore, we also introduce an ensemble method to enrich the input data by creating synthetic cell compositions. To create a synthetic slide, we randomly selected $30\%$ cells from one slide and $1\%$ cells from all the other slides in the same class as all the cells in the synthetic slide. Then, the synthetic slide will represent the main pattern of one observed slide but also introduce other variations. We created 100 synthetic slides for each class using this approach, which is indicated as ''Ensemble'' in Table~\ref{tab:classification_methods}.
The simple ANN we used is $\text{fc}(8) +\text{ReLU} +\text{fc}(1)$, where $n$ in $\text{fc}(n)$ means the number of output channels. 
\new{To evaluate the methods, we then perform 100 random training/testing splits of our dataset on the patient-level and train all models on every split. The mean results and the corresponding confidence intervals are summarized in Table~\ref{tab:classification_methods}.}

Logistic regression has the lowest testing accuracy ($75.14\%$) and the rule set achieves the highest performance ($87.50\%$), which is comparable to SVM ($82.07\%$) and ANN ($83.71\%$). Creating the embedding from distinct \ac{GEP} classes results in similar accuracy of the rule set algorithm ($87.50\%$ v.s. $84.33\%$). 
As in the previous comparisons to black box models, the rule set approach has the added benefit of being interpretable. Logistic regression and SVM models suffer in this regard due to the high dimensionality of the input representation (78).
Dimensionality reduction techniques, \eg principle component analysis (PCA), exist but are not applicable here because the number of input variables (78) is larger than the number of training samples (64). Finally, all models reach higher accuracy with the ensemble except SVM. 

Due to the fact that our segmentation model is not perfect, we also evaluate the rule set model for different segmentation results. During early training, the segmentation model will first identify the most clear cancer cells, but along with plenty of false positives. As the optimization progresses, fewer cancer cells are segmented but much fewer false positives occur. 
The accuracy of the rule set algorithm for segmentation results after 2000, 3000 and 4000 training iterations is $77.23\pm 10.98\%$, $84.64 \pm 10.46\%$ and $87.50\pm 9.56\%$, which suggests that the algorithm favors the output of a highly specific cell segmentation algorithm. 


\section{Discussion}
\new{
Our overall system utilizes the segmentation features to generate the embedding and classify \ac{UM} subtypes based on slide-level cell composition. We assume that cell composition will be different across the two \ac{GEP} classes, which will result in different cell density chart representations of slides from the two subtypes that can then be distinguished using an interpretable rule-based algorithm. This hypothesis is supported by our experiments. 
One aspect of the current approach is that we do not currently interpret the embedded features themselves, e.\,g., by classifying cell types, so that regions in the density chart do not immediately carry semantic information. 
This circumstance may limit the interpretability of our tool for non-subject matter experts, however, we emphasize that the tool is designed with pathologists as primary user group in mind who possess substantial domain expertise. While the algorithm does not currently identify specific cell types during embedding, pathologists are domain experts and will be able to explore and contextualize cell appearance in different embedding regions using the graphical user interface (Figure~\ref{fig:GUI}). Doing so as part of a training period may allow pathologists to understand and identify the major cell types in specific embedding regions, linking pie chart sectors to semantic cell types. Experience in observing how \ac{GEP} class 1 and class 2 slides behave in the pie chart embedding space combined with the above training may further add to the interpretability of the model. Future work will investigate how this paradigm compares to other approaches and black box models in building trust and confidence in the user group.}

\new{
In the current form, the circular space is evenly partitioned into 12 parts. However, this partitioning process could be further guided by other semantic information, \eg clustering of specific cell types. If cell type annotations are available, the segmentation network could also output cell types for every extracted cells. We would expect to see clusters of cells types in the embedding space and the partitioning process could be further guided by the cell type clusters. However, such information is not available in our dataset and it is also unclear whether such approach would prove beneficial.
}

\new{
The boundary defined by the interpretable rule set could also be used as a criterion for user trust calibration. If a slide maps close to the boundary, a little variation could change the prediction result. Thus, the prediction of that slide may be perceived as less reliable. We will investigate in future work whether proximity to the decision boundary indeed correlates with prediction performance, and more importantly, other clinical outcome measures such as survival. 
}

\section{Conclusion}
We have presented an automated yet interpretable system for \ac{UM} subtyping from fine needle aspiration cytology images that does not compromise performance compared to conventional deep black box models. In future work, we will study how our interpretable model affects treatment decisions and user trust, as a next step to realize the huge potential of image-based tests for \ac{UM} subtyping.

\noindent\textbf{Acknowledgement}: We gratefully acknowledge funding from the Emerson Collective Cancer Research Fund and internal funds provided by the Wilmer Eye Institute and the Malone Center for Engineering in Healthcare at Johns Hopkins University.  

\newpage
\bibliographystyle{icml2021}
\bibliography{references}

\begin{thebibliography}{33}
\providecommand{\natexlab}[1]{#1}
\providecommand{\url}[1]{\texttt{#1}}
\expandafter\ifx\csname urlstyle\endcsname\relax
  \providecommand{\doi}[1]{doi: #1}\else
  \providecommand{\doi}{doi: \begingroup \urlstyle{rm}\Url}\fi

\bibitem[Achanta et~al.(2010)Achanta, Shaji, Smith, Lucchi, Fua, and
  Süsstrunk]{SLIC}
Achanta, R., Shaji, A., Smith, K., Lucchi, A., Fua, P., and Süsstrunk, S.
\newblock Slic superpixels, 2010.

\bibitem[Bolya et~al.(2019)Bolya, Zhou, Xiao, and Lee]{YOLACT}
Bolya, D., Zhou, C., Xiao, F., and Lee, Y.~J.
\newblock Yolact: Real-time instance segmentation.
\newblock In \emph{Proceedings of the IEEE/CVF International Conference on
  Computer Vision (ICCV)}, October 2019.

\bibitem[Campanella et~al.(2019)Campanella, Hanna, Geneslaw, Miraflor, Silva,
  Busam, Brogi, Reuter, Klimstra, and Fuchs]{campanella2019clinical}
Campanella, G., Hanna, M.~G., Geneslaw, L., Miraflor, A., Silva, V. W.~K.,
  Busam, K.~J., Brogi, E., Reuter, V.~E., Klimstra, D.~S., and Fuchs, T.~J.
\newblock Clinical-grade computational pathology using weakly supervised deep
  learning on whole slide images.
\newblock \emph{Nature medicine}, 25\penalty0 (8):\penalty0 1301--1309, 2019.

\bibitem[Chen et~al.(2020)Chen, Liu, Correa, and Unberath]{chen2020interactive}
Chen, H., Liu, T.~A., Correa, Z., and Unberath, M.
\newblock An interactive approach to region of interest selection in cytologic
  analysis of uveal melanoma based on unsupervised clustering.
\newblock In \emph{International Workshop on Ophthalmic Medical Image
  Analysis}, pp.\  114--124. Springer, 2020.

\bibitem[Chikontwe et~al.(2020)Chikontwe, Kim, Nam, Go, and
  Park]{chikontwe2020multiple}
Chikontwe, P., Kim, M., Nam, S.~J., Go, H., and Park, S.~H.
\newblock Multiple instance learning with center embeddings for histopathology
  classification.
\newblock In \emph{International Conference on Medical Image Computing and
  Computer-Assisted Intervention}, pp.\  519--528. Springer, 2020.

\bibitem[Coudray et~al.(2018)Coudray, Ocampo, Sakellaropoulos, Narula, Snuderl,
  Feny{\"o}, Moreira, Razavian, and Tsirigos]{coudray2018classification}
Coudray, N., Ocampo, P.~S., Sakellaropoulos, T., Narula, N., Snuderl, M.,
  Feny{\"o}, D., Moreira, A.~L., Razavian, N., and Tsirigos, A.
\newblock Classification and mutation prediction from non--small cell lung
  cancer histopathology images using deep learning.
\newblock \emph{Nature medicine}, 24\penalty0 (10):\penalty0 1559--1567, 2018.

\bibitem[Cowan(2010)]{cowan2010magical}
Cowan, N.
\newblock The magical mystery four: How is working memory capacity limited, and
  why?
\newblock \emph{Current directions in psychological science}, 19\penalty0
  (1):\penalty0 51--57, 2010.

\bibitem[Gaube et~al.(2021)Gaube, Suresh, Raue, Merritt, Berkowitz, Lermer,
  Coughlin, Guttag, Colak, and Ghassemi]{gaube2021ai}
Gaube, S., Suresh, H., Raue, M., Merritt, A., Berkowitz, S.~J., Lermer, E.,
  Coughlin, J.~F., Guttag, J.~V., Colak, E., and Ghassemi, M.
\newblock Do as ai say: susceptibility in deployment of clinical decision-aids.
\newblock \emph{npj Digital Medicine}, 4\penalty0 (1):\penalty0 1--8, 2021.

\bibitem[Grossmann et~al.(2017)Grossmann, Stringfield, El-Hachem, Bui,
  Velazquez, Parmar, Leijenaar, Haibe-Kains, Lambin, Gillies, and
  Aerts]{Grossmann2017DefiningTB}
Grossmann, P., Stringfield, O., El-Hachem, N., Bui, M.~M., Velazquez, E.~R.,
  Parmar, C., Leijenaar, R. T.~H., Haibe-Kains, B., Lambin, P., Gillies, R.~J.,
  and Aerts, H.~J.
\newblock Defining the biological basis of radiomic phenotypes in lung cancer.
\newblock In \emph{eLife}, 2017.

\bibitem[He et~al.(2015)He, Zhang, Ren, and Sun]{He2015}
He, K., Zhang, X., Ren, S., and Sun, J.
\newblock Deep residual learning for image recognition.
\newblock \emph{arXiv preprint arXiv:1512.03385}, 2015.

\bibitem[He et~al.(2017)He, Gkioxari, Doll{\'a}r, and Girshick]{he2017mask}
He, K., Gkioxari, G., Doll{\'a}r, P., and Girshick, R.
\newblock Mask r-cnn.
\newblock In \emph{Proceedings of the IEEE international conference on computer
  vision}, pp.\  2961--2969, 2017.

\bibitem[Hegde et~al.(2019)Hegde, Hipp, Liu, Emmert-Buck, Reif, Smilkov, Terry,
  Cai, Amin, Mermel, et~al.]{hegde2019similar}
Hegde, N., Hipp, J.~D., Liu, Y., Emmert-Buck, M., Reif, E., Smilkov, D., Terry,
  M., Cai, C.~J., Amin, M.~B., Mermel, C.~H., et~al.
\newblock Similar image search for histopathology: Smily.
\newblock \emph{NPJ digital medicine}, 2\penalty0 (1):\penalty0 1--9, 2019.

\bibitem[Hou et~al.(2016)Hou, Samaras, Kurc, Gao, Davis, and
  Saltz]{hou2016patch}
Hou, L., Samaras, D., Kurc, T.~M., Gao, Y., Davis, J.~E., and Saltz, J.~H.
\newblock Patch-based convolutional neural network for whole slide tissue image
  classification.
\newblock In \emph{Proceedings of the IEEE conference on computer vision and
  pattern recognition}, pp.\  2424--2433, 2016.

\bibitem[Kim()]{fastSLIC}
Kim, A.
\newblock Fastslic: Optimized slic superpixel.
\newblock URL \url{https://github.com/Algy/fast-slic}.

\bibitem[Kingma \& Ba(2014)Kingma and Ba]{kingma2014adam}
Kingma, D.~P. and Ba, J.
\newblock Adam: A method for stochastic optimization.
\newblock \emph{arXiv preprint arXiv:1412.6980}, 2014.

\bibitem[Kolmogorov(1933)]{kolmogorov1933sulla}
Kolmogorov, A.
\newblock Sulla determinazione empirica di una lgge di distribuzione.
\newblock \emph{Inst. Ital. Attuari, Giorn.}, 4:\penalty0 83--91, 1933.

\bibitem[Liu et~al.(2020)Liu, Zhu, Chen, Arevalo, Hui, Paul, Wei, Unberath, and
  Correa]{liu2020gene}
Liu, T.~A., Zhu, H., Chen, H., Arevalo, J.~F., Hui, F.~K., Paul, H.~Y., Wei,
  J., Unberath, M., and Correa, Z.~M.
\newblock Gene expression profile prediction in uveal melanoma using deep
  learning: A pilot study for the development of an alternative survival
  prediction tool.
\newblock \emph{Ophthalmology. Retina}, 4\penalty0 (12):\penalty0 1213--1215,
  2020.

\bibitem[Liu et~al.(2017)Liu, Gadepalli, Norouzi, Dahl, Kohlberger, Boyko,
  Venugopalan, Timofeev, Nelson, Corrado, et~al.]{liu2017detecting}
Liu, Y., Gadepalli, K., Norouzi, M., Dahl, G.~E., Kohlberger, T., Boyko, A.,
  Venugopalan, S., Timofeev, A., Nelson, P.~Q., Corrado, G.~S., et~al.
\newblock Detecting cancer metastases on gigapixel pathology images.
\newblock \emph{arXiv preprint arXiv:1703.02442}, 2017.

\bibitem[McInnes et~al.(2020)McInnes, Healy, and Melville]{UMAP}
McInnes, L., Healy, J., and Melville, J.
\newblock Umap: Uniform manifold approximation and projection for dimension
  reduction, 2020.

\bibitem[Miller(2020)]{miller202011}
Miller, G.~A.
\newblock \emph{11. The magical number seven, plus-or-minus two or some limits
  on our capacity for processing information}.
\newblock University of California Press, 2020.

\bibitem[Murdoch et~al.(2019)Murdoch, Singh, Kumbier, Abbasi-Asl, and
  Yu]{Murdoch2019}
Murdoch, W.~J., Singh, C., Kumbier, K., Abbasi-Asl, R., and Yu, B.
\newblock Definitions, methods, and applications in interpretable machine
  learning.
\newblock \emph{Proceedings of the National Academy of Sciences}, 116\penalty0
  (44):\penalty0 22071--22080, 2019.
\newblock ISSN 0027-8424.
\newblock \doi{10.1073/pnas.1900654116}.
\newblock URL \url{https://www.pnas.org/content/116/44/22071}.

\bibitem[Nourani et~al.(2020)Nourani, King, and Ragan]{nourani2020role}
Nourani, M., King, J., and Ragan, E.
\newblock The role of domain expertise in user trust and the impact of first
  impressions with intelligent systems.
\newblock In \emph{Proceedings of the AAAI Conference on Human Computation and
  Crowdsourcing}, volume~8, pp.\  112--121, 2020.

\bibitem[Onken et~al.(2004)Onken, Worley, Ehlers, and Harbour]{onken2004gene}
Onken, M.~D., Worley, L.~A., Ehlers, J.~P., and Harbour, J.~W.
\newblock Gene expression profiling in uveal melanoma reveals two molecular
  classes and predicts metastatic death.
\newblock \emph{Cancer research}, 64\penalty0 (20):\penalty0 7205--7209, 2004.

\bibitem[Pearson(1900)]{pearson1900x}
Pearson, K.
\newblock X. on the criterion that a given system of deviations from the
  probable in the case of a correlated system of variables is such that it can
  be reasonably supposed to have arisen from random sampling.
\newblock \emph{The London, Edinburgh, and Dublin Philosophical Magazine and
  Journal of Science}, 50\penalty0 (302):\penalty0 157--175, 1900.

\bibitem[Peng et~al.(2019)Peng, Boxberg, Weichert, Navab, and
  Marr]{peng2019multi}
Peng, T., Boxberg, M., Weichert, W., Navab, N., and Marr, C.
\newblock Multi-task learning of a deep k-nearest neighbour network for
  histopathological image classification and retrieval.
\newblock In \emph{International Conference on Medical Image Computing and
  Computer-Assisted Intervention}, pp.\  676--684. Springer, 2019.

\bibitem[Pirovano et~al.(2020)Pirovano, Heuberger, Berlemont, Ladjal, and
  Bloch]{pirovano2020improving}
Pirovano, A., Heuberger, H., Berlemont, S., Ladjal, S., and Bloch, I.
\newblock Improving interpretability for computer-aided diagnosis tools on
  whole slide imaging with multiple instance learning and gradient-based
  explanations.
\newblock In \emph{Interpretable and Annotation-Efficient Learning for Medical
  Image Computing}, pp.\  43--53. Springer, 2020.

\bibitem[Rudin(2019)]{rudin2019stop}
Rudin, C.
\newblock Stop explaining black box machine learning models for high stakes
  decisions and use interpretable models instead.
\newblock \emph{Nature Machine Intelligence}, 1\penalty0 (5):\penalty0
  206--215, 2019.

\bibitem[Schopper \& Correa(2016)Schopper and Correa]{schopper2016clinical}
Schopper, V.~J. and Correa, Z.~M.
\newblock Clinical application of genetic testing for posterior uveal melanoma.
\newblock \emph{International journal of retina and vitreous}, 2\penalty0
  (1):\penalty0 1--6, 2016.

\bibitem[Singh et~al.(2011)Singh, Turell, and Topham]{SINGH20111881}
Singh, A.~D., Turell, M.~E., and Topham, A.~K.
\newblock Uveal melanoma: Trends in incidence, treatment, and survival.
\newblock \emph{Ophthalmology}, 118\penalty0 (9):\penalty0 1881 -- 1885, 2011.
\newblock ISSN 0161-6420.

\bibitem[Wang et~al.(2017)Wang, Rudin, Doshi-Velez, Liu, Klampfl, and
  MacNeille]{wang2017bayesian}
Wang, T., Rudin, C., Doshi-Velez, F., Liu, Y., Klampfl, E., and MacNeille, P.
\newblock A bayesian framework for learning rule sets for interpretable
  classification.
\newblock \emph{The Journal of Machine Learning Research}, 18\penalty0
  (1):\penalty0 2357--2393, 2017.

\bibitem[Worley et~al.(2007)Worley, Onken, Person, Robirds, Branson, Char,
  Perry, and Harbour]{worley2007transcriptomic}
Worley, L.~A., Onken, M.~D., Person, E., Robirds, D., Branson, J., Char, D.~H.,
  Perry, A., and Harbour, J.~W.
\newblock Transcriptomic versus chromosomal prognostic markers and clinical
  outcome in uveal melanoma.
\newblock \emph{Clinical Cancer Research}, 13\penalty0 (5):\penalty0
  1466--1471, 2007.

\bibitem[Xu et~al.(2017)Xu, Jia, Wang, Ai, Zhang, Lai, Eric, and
  Chang]{xu2017large}
Xu, Y., Jia, Z., Wang, L.-B., Ai, Y., Zhang, F., Lai, M., Eric, I., and Chang,
  C.
\newblock Large scale tissue histopathology image classification, segmentation,
  and visualization via deep convolutional activation features.
\newblock \emph{BMC bioinformatics}, 18\penalty0 (1):\penalty0 1--17, 2017.

\bibitem[Zhang et~al.(2020)Zhang, Kalirai, Acha-Sagredo, Yang, Zheng, and
  Coupland]{zhang2020piloting}
Zhang, H., Kalirai, H., Acha-Sagredo, A., Yang, X., Zheng, Y., and Coupland,
  S.~E.
\newblock Piloting a deep learning model for predicting nuclear bap1
  immunohistochemical expression of uveal melanoma from hematoxylin-and-eosin
  sections.
\newblock \emph{Translational Vision Science \& Technology}, 9\penalty0
  (2):\penalty0 50--50, 2020.

\end{thebibliography}
\end{document}